\RequirePackage{fix-cm}
\documentclass[letterpaper, 10 pt, conference]{ieeeconf}  
\usepackage{caption}
\usepackage{footnote}
\usepackage{url}  
\usepackage{enumerate}
\usepackage{graphicx,color}
\usepackage{amsmath,amssymb} 
\usepackage{bm}
\usepackage{epstopdf}
\usepackage{wrapfig}
\usepackage{multirow}
\usepackage{booktabs}
\usepackage{tabularx}
\usepackage{threeparttable}
\usepackage{algorithm}
\usepackage{algpseudocode}
\usepackage{cite}
\algtext*{EndWhile}
\algtext*{EndIf}
\algtext*{EndFor}

\IEEEoverridecommandlockouts                              
\title{\LARGE \textbf{Densely Supervised Grasp Detector (DSGD)}}
\author{Umar Asif, Jianbin Tang, and Stefan Harrer
\thanks{Umar Asif, Jianbin Tang, and Stefan Harrer are with IBM Research Australia. Email:\{umarasif, jbtang, sharrer\}@au1.ibm.com.}%
\thanks{}}
\begin{document}
\maketitle
\thispagestyle{empty}
\pagestyle{empty}
%
\begin{abstract}
This paper presents \textit{Densely Supervised Grasp Detector (DSGD)}, a deep learning framework which combines CNN structures with layer-wise feature fusion and produces grasps and their confidence scores at different levels of the image hierarchy (i.e., global-, region-, and pixel-levels).
Specifically, at the global-level, DSGD uses the entire image information to predict a grasp. At the region-level, DSGD uses a region proposal network to identify salient regions in the image and predicts a grasp for each salient region. At the pixel-level, DSGD uses a fully convolutional network and predicts a grasp and its confidence at every pixel. 
During inference, DSGD selects the most confident grasp as the output. This selection from hierarchically generated grasp candidates overcomes limitations of the individual models. 
DSGD outperforms state-of-the-art methods on the Cornell grasp dataset in terms of grasp accuracy. 
Evaluation on a multi-object dataset and real-world robotic grasping experiments show that DSGD produces highly stable grasps on a set of unseen objects in new environments. It achieves 97\% grasp detection accuracy and 90\% robotic grasping success rate with real-time inference speed. 
\end{abstract}
%
\section{Introduction}\label{introduction}
Grasp detection is a crucial task in robotic grasping because errors in this stage affect grasp planning and execution. A major challenge in grasp detection is generalization to unseen objects in the real-world.  
Recent advancements in deep learning have produced Convolutional Neural Network (CNN) based grasp detection methods which achieve higher grasp detection accuracy compared to hand-crafted features. 
Methods such as \cite{lenz2015deep,redmon2015real,asif2017rgb,asifbmvc2018} focused on learning grasps in a global-context (i.e., the model predicts one grasp considering the whole input image), through regression-based approaches (which directly regress the grasp parameters defined by the location, width, height, and orientation of a 2D rectangle in image space). 
Other methods such as \cite{pinto2016supersizing} focused on learning grasps at patch-level by extracting patches (of different sizes) from the image and predicting a grasp for each patch.
Recently, methods such as \cite{morrison2018closing,zeng2017robotic} used auto-encoders to learn grasp parameters at each pixel in the image. They showed that one-to-one mapping (of image data to ground truth grasps) at the pixel-level can effectively be learnt using small CNN structures to achieve fast inference speed.
These studies show that grasp detection performance is strongly influenced by three main factors: \textbf{i)} The choice of the CNN structure used for feature learning, \textbf{ii)} the objective function used to learn grasp representations, and \textbf{iii)} the image hierarchical context at which grasps are learnt (e.g., global or local). 
\\
\indent
In this work, we explore the advantages of combining multiple global and local grasp detectors and a mechanism to select the best grasp out of the ensemble. 
We also explore the benefits of learning grasp parameters using a combination of regression and classification objective functions. Finally, we explore different CNN structures as base networks to identify the best performing architecture in terms of grasp detection accuracy.
The main contributions of this paper are summarized below:
\begin{enumerate}[$\hspace{5pt}$1)]
\item
We present \textit{Densely Supervised Grasp Detector} (DSGD), an ensemble of multiple CNN structures which generate grasps and their confidence scores at different levels of image hierarchy (i.e., global-level, region-level, and pixel-level). 
%
%
\item
We propose a region-based grasp network, which learns to extract salient parts (e.g., handles or boundaries) from the input image, and uses the information about these parts to learn class-specific grasps (i.e., each grasp is associated with a probability with respect to a graspable class and a non-graspable class). 
\item
We perform an ablation study of our DSGD by varying its critical parameters and present a grasp detector that achieves real-time speed and high grasp accuracy.
\item
We demonstrate the robustness of DSGD for producing stable grasps for unseen objects in real-world environments using a multi-object dataset and robotic grasping experiments.
\end{enumerate}
\section{Related Work}\label{related_work}
In the context of deep learning based grasp detection, methods such as \cite{saxena2008robotic,jiang2011efficient,lenz2015deep} trained sliding window based grasp detectors. However, their high inference times limit their application for real-time systems. Other methods such as \cite{mahler2016dex,mahler2017dex,johns2016deep} reduced inference time by processing a discrete set of grasp candidates, but these methods ignore some potential grasps.
Alternatively, methods such as \cite{redmon2015real,kumra2016robotic,guo2017hybrid} proposed end-to-end CNN-based approaches which regress a single grasp for an input image. However, these methods tend to produce average grasps which are invalid for certain symmetric objects \cite{redmon2015real}.
Recently, methods such as \cite{morrison2018closing,zeng2017robotic,zeng2018robotic,myers2015affordance,varley2015generating} used auto-encoders to generate grasp poses at every pixel. They demonstrated higher grasp accuracy compared to the global methods.
Another stream of work focused on learning mapping between images of objects and robot motion parameters using reinforcement learning, where the robot iteratively refines grasp poses through real-world experiments. 
In this context, the method of \cite{pinto2016supersizing} learned visual-motor control by performing more than 40k grasping trials on a real robot. The method of \cite{levine2016learning} learned image-to-gripper pose mapping using over 800k motor commands generated by 14 robotic arms performing grasping trials for over 2 months. 
\\
\indent
In this paper, we present a grasp detector which has several key differences from the current grasp detection methods. \textbf{First}, our detector  generates multiple global and local grasp candidates and selects the grasp with the highest quality. This allows our detector to effectively recover from the errors of the individual global or local models. 
\textbf{Second}, we introduce a region-based grasp network which learns grasps using information about salient parts of objects (e.g., handles, extrusions, or boundaries), and produces more accurate grasps compared to global \cite{kumra2016robotic} or local detectors \cite{pinto2016supersizing}. 
\textbf{Finally}, we use layer-wise dense feature fusion \cite{huang2017densely} within the CNN structures. This maximizes variation in the information flow across the networks and produces better image-to-grasp mappings compared to the models of \cite{redmon2015real,morrison2018closing}. 
\section{Problem Formulation}\label{problem_description}
\begin{figure*}[t]
  \begin{center}
    \includegraphics[trim=0.2cm 0.2cm 0.2cm 0.0cm,clip,width=1.0\linewidth,keepaspectratio]{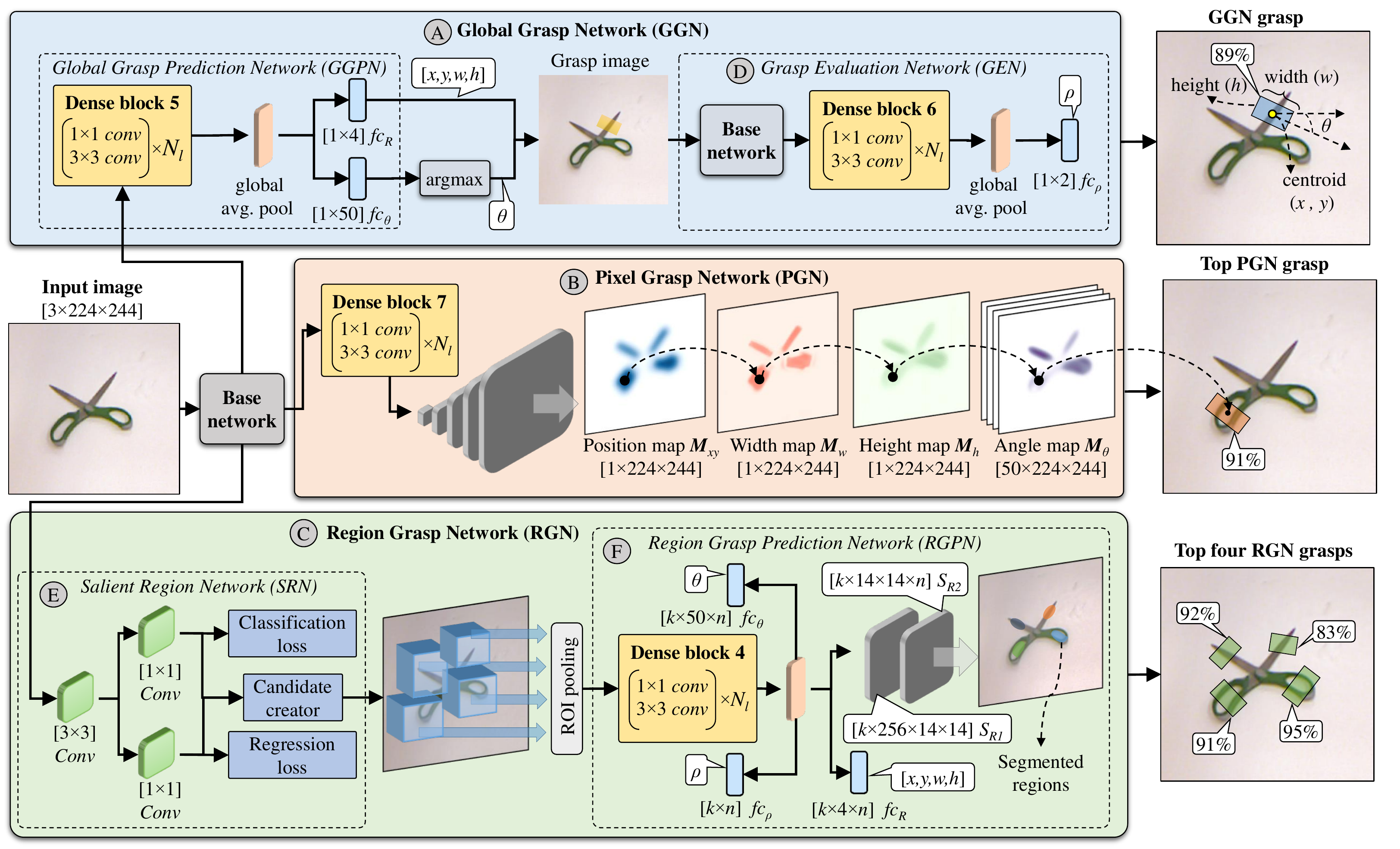}
  \vspace{-15pt}
  \caption{Overview of our DSGD architecture. Given an image as input, DSGD uses a base network to extract features which are fed into a Global Grasp Network (A), a Pixel Grasp Network (B), and a Region Grasp network (C), to produce grasp candidates. The global model produces a single grasp per image and uses an independent Grasp Evaluation Network (D) to produce grasp confidence. The pixel-level model uses a fully convolutional network and produces grasps at every pixel. The region-level model uses a Salient Region Network (E) to extract salient parts of the image and uses information about these parts to produce grasps. During inference, DSGD switches between the GGN, the PGN, and the RGN models based on their confidence scores.}
  \vspace{-10pt}
  \label{fig_network}
  \end{center}
\end{figure*}

Given an image of an object as input, the goal is to generate grasps at different image hierarchical levels (i.e., global-, region- and pixel-levels), and select the most confident grasp as the output. We define the global grasp by a 2D oriented rectangle on the target object in the image space. It is given by:
\begin{equation}
\mathcal{G}_g=[x_g,y_g,w_g,h_g,\theta_g,\rho_g],
\end{equation}
where $x_g$ and $y_g$ represent the centroid of the rectangle. The terms $w_g$ and $h_g$ represent the width and the height of the rectangle. The term $\theta_g$ represents the angle of the rectangle with respect to x-axis. The term $\rho_g$ is \textit{grasp confidence} and represents the quality of a grasp.
Our region-level grasp is defined by a class-specific representation, where the parameters of the rectangle are associated with $n$ classes (a graspable class: $n=1$, and a non-graspable class: $n=0$). It is given by: 
\begin{equation}
\mathcal{G}_r=[x_r^{n},y_r^{n},w_r^{n},h_r^{n},\theta_r^{n},\rho_r^{n}], n\in[0,1].
\end{equation}
Our pixel-level grasp is defined as:
\begin{equation}
\mathcal{G}_p=[\boldsymbol{M}_{xy},\boldsymbol{M}_w,\boldsymbol{M}_h,\boldsymbol{M}_\theta]\in\mathbb{R}^{s\times W\times H},
\end{equation}
where $\boldsymbol{M}_{xy}$, $\boldsymbol{M}_w$, $\boldsymbol{M}_h$, and $\boldsymbol{M}_\theta$ represent $\mathbb{R}^{s\times W\times H}-$dimensional heatmaps\footnote{$s=1$ for $\boldsymbol{M}_{xy}$, $\boldsymbol{M}_w$, and $\boldsymbol{M}_h$, and $s=N_{\theta}$ for $\boldsymbol{M}_\theta$.}, which encode the position, width, height, and orientation of grasps at every pixel of the image, respectively. The terms $W$ and $H$ represent the width and the height of the input image respectively.
We learn the grasp representations ($\mathcal{G}_g$, $\mathcal{G}_r$, and $\mathcal{G}_p$) using joint regression-classification based objective functions. Specifically, we learn the position, width, and the height parameters using a Mean Squared Loss, and learn the orientation parameter using a Cross Entropy Loss with respect to $N_{\theta}=50$ classes (angular-bins).
\section{The Proposed DSGD (Fig. \ref{fig_network})}\label{proposed_model}
\begin{figure}[t]
  \begin{center}
    \includegraphics[trim=0.0cm 0.0cm 0.0cm 0.0cm,clip,width=1.0\linewidth,keepaspectratio]{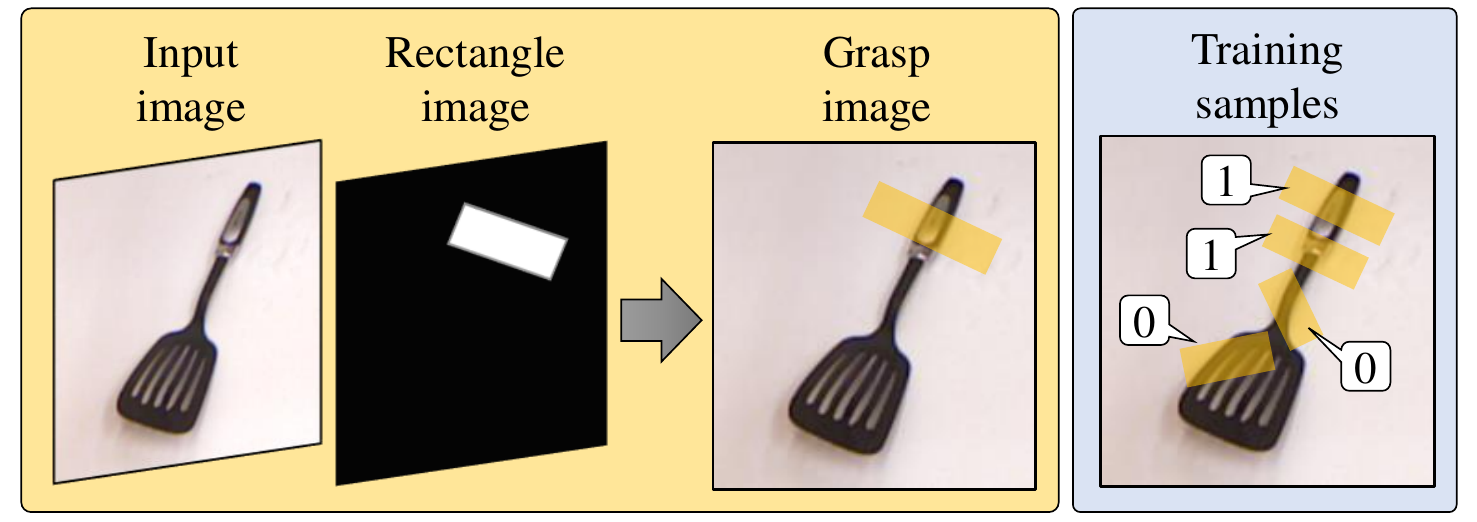}
  \vspace{-18pt}
  \caption{Left: A grasp image is generated by replacing the blue channel of the input image with a binary rectangle image produced from a grasp pose. Right: Our Grasp Evaluation Network is trained using grasp images labelled in terms of valid (1) and invalid (0) grasp rectangles.} 
  \vspace{-12pt}
  \label{fig_gen}
  \end{center}
\end{figure}
\begin{figure*}[t]
  \begin{center}
    \includegraphics[trim=0.0cm 0.0cm 0.0cm 0.0cm,clip,width=1.0\linewidth,keepaspectratio]{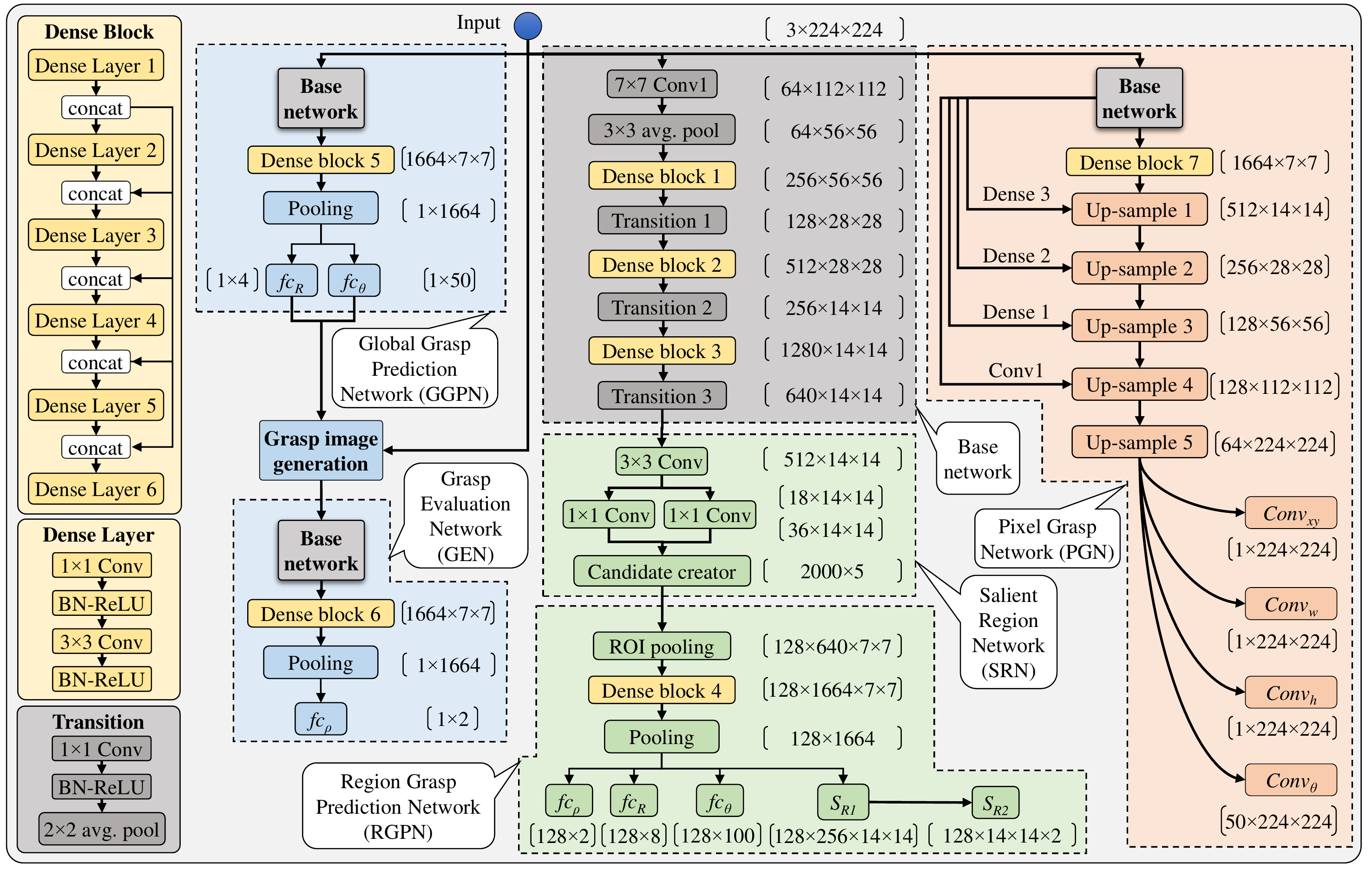}
  \vspace{-15pt}
  \caption{Detailed architecture of our DSGD with a DenseNet \cite{huang2017densely} as its base network.
  }
  \vspace{-12pt}
  \label{fig_network2}
  \end{center}
\end{figure*}
Our DSGD is composed of four main modules as shown in Fig. \ref{fig_network}: a base network for feature extraction, a Global Grasp Network (GGN) for producing a grasp at the image-level, a Region Grasp Network (RGN) for producing grasps using salient parts of the image, and a Pixel Grasp Network (PGN) for generating grasps at each image pixel. 
In the following, we describe in detail the various modules of DSGD.
\subsection{Base Network}\label{base_network}
The purpose of the base network is to act as a feature extractor. We extract features from the intermediate layers of a CNN such as DenseNets \cite{huang2017densely}, and use the features to learn grasp representations at different hierarchical levels. 
%
%
%
The basic building block of DenseNets \cite{huang2017densely} is a \textit{Dense block}: bottleneck convolutions interconnected through dense connections.
Specifically, a dense block consists of $N_{l}$ number of layers termed \textit{Dense Layers} which share information from all the preceding layers connected to the current layer through skip connections \cite{huang2017densely}. Fig. \ref{fig_network2} shows the structure of a dense block with $N_{l}=6$ dense layers. Each dense layer consists of $1\times1$ and $3\times3$ convolutions followed by Batch Normalization \cite{ioffe2015batch} and a Rectified Linear Unit (ReLU).
The output of the $l^{th}$ dense layer ($\mathcal{X}_l$) in a dense block can be written as: 
\begin{equation}
\mathcal{X}_l=[\mathcal{X}_0,...,\mathcal{X}_{l-1}],
\end{equation}
where $[\cdot\cdot\cdot]$ represents concatenation of the features produced by the layers $0,...,l-1$.
\subsection{Global Grasp Network (GGN)}\label{global_grasp_network}
Our GGN structure is composed of two sub-networks as shown in Fig. \ref{fig_network}-A: a Global Grasp Prediction Network (GGPN) for generating grasp pose ($[x_g,y_g,w_g,h_g,\theta_g]$) and a Grasp Evaluation Network (GEN) for predicting grasp confidence ($\rho_g$). 
The GGPN structure is composed of a dense block, an averaging operation, a $4-$dimensional fully connected layer for predicting the parameters $[x_g,y_g,w_g,h_g]$, and a $50-$dimensional fully connected layer for predicting $\theta_g$. 
The GEN structure is similar to GGPN except that GEN has a single $2-$dimensional fully connected layer for predicting $\rho_g$.
The input to GEN is a grasp image which is produced by replacing the \textit{Blue} channel of the input image with a binary rectangle image generated from the output of GGPN as shown in Fig. \ref{fig_gen}. 
\\
\indent
Let $R_{g_i}=[x_{g_i},y_{g_i},w_{g_i},h_{g_i}]$, $\theta_{g_i}$ and $\rho_{g_i}$ denote the predicted values of a global grasp for the $i^{th}$ image. We define the loss of the GGPN and the GEN models over $\boldsymbol{K}$ images as:
\begin{equation}\label{loss_ggpn}
L_{ggpn}=\sum_{i\in \boldsymbol{K}}\left((1-\lambda_1)L_{reg}(R_{g_i},R_{g_i}^{*})+\lambda_{1}L_{cls}(\theta_{g_i},\theta_{g_i}^{*})\right),
\end{equation}
\begin{equation}\label{loss_gen}
L_{gen}=\sum_{i\in \boldsymbol{K}}L_{cls}(\rho_{g_i},\rho_{g_i}^{*}),
\end{equation}
where $R_{g_i}^{*}$, $\theta^{*}_{g_i}$, and $\rho^{*}_{g_i}$ represent the ground-truths. The term $L_{reg}$ is a regression loss defined as:
\begin{equation}\label{loss_reg}
L_{reg}({R},R^{*})={||{R}-R^{*}||}/{||R^{*}||_{2}}.
\end{equation}
The term $L_{cls}$ is a classification loss defined as:
\begin{equation}\label{loss_cls}
L_{cls}(\boldsymbol{\textup{x}},c)=-\sum_{c=1}^{N_{\theta}}\mathcal{Y}_{\textup{x},c}\log(p_{\textup{x},c}),
\end{equation}
where $\mathcal{Y}$ is a binary indicator if class label $c$ is the correct classification for observation $\textup{x}$, and $p$ is the predicted probability of observation $\textup{x}$ of class $c$.
\subsection{Region Grasp Network (RGN)}\label{region_grasp_network}
The RGN structure is composed of two sub-networks as shown in Fig. \ref{fig_network}-C: a Salient Region Network (SRN) for extracting salient parts of image, and a region grasp prediction network (RGPN) for predicting a grasp for each candidate salient part.
\subsubsection{Salient Region Network (SRN):}\label{salient_region_network}
Here, we use the features extracted from the base network to generate regions defined by the location ($x_{sr},y_{sr}$), width ($w_{sr}$), height ($h_{sr}$), and confidence ($\rho_{sr}$) of non-oriented rectangles which encompass salient parts of the image (e.g., handles, extrusions, or boundaries). 
For this, we first generate a fixed number of rectangles using the Region of Interest (ROI) method of \cite{he2017mask}. Next, we use the features from the base network and optimize a Mean Squared Loss on the rectangle coordinates and a Cross Entropy Loss on the rectangle confidence scores.
\\
\indent
Let $T_i=[x_{sr},y_{sr},w_{sr},h_{sr}]$ denote the parameters of the $i^{th}$ predicted rectangle, and $\rho_{sr_i}$ denote its probability whether it belongs to a graspable region or a non-graspable region. The loss of SRN over $\boldsymbol{I}$ proposals is given by:
\begin{equation}\label{loss_srn}
L_{srn}=\sum_{i\in \boldsymbol{I}}\left((1-\lambda_2)L_{reg}(T_{i},T_{i}^{*})+\lambda_2L_{cls}(\rho_{sr_i},\rho_{sr_i}^{*})\right),
\end{equation}
where $\rho_{sr_i}^{*}=0$ for a non-graspable region and $\rho_{sr_i}^{*}=1$ for a graspable region. The term $T_{i}^{*}$ represents the ground truth candidate corresponding to $\rho_{sr_i}^{*}$.
\subsubsection{Region Grasp Prediction Network (RGPN):}\label{region_grasp_prediction}
Here, we produce grasp poses for the salient regions predicted by SRN ($k=128$ in our implementation). For this, we crop features from the output feature maps of the base network using the Region of Interest (ROI) pooling method of \cite{he2017mask}. The cropped features are then fed to \textit{Dense block 4} which produces feature maps of $k\times1664\times7\times7-$dimensions as shown in Fig. \ref{fig_network2}. These feature maps are then squeezed to $k\times1664-$dimensions through a global average pooling, and fed to three fully connected layers $fc_{R}\in\mathbb{R}^{k\times4\times n}$, $fc_{\theta}\in\mathbb{R}^{k\times50\times n}$, and $fc_{\rho}\in\mathbb{R}^{k\times2}$ which produce class-specific grasps. RGPN also has a segmentation branch (with two upsampling layers $S_{R1}\in\mathbb{R}^{k\times256\times14\times14}$ and $S_{R2}\in\mathbb{R}^{k\times14\times14\times n}$), which produces a segmentation mask for each salient region as shown in Fig. \ref{fig_network}-F. 
\\
\indent
Let $R_{r_i}=[x_{r_i},y_{r_i},w_{r_i},h_{r_i}]$, $\theta_{r_i}$, and $\rho_{r_i}$ denote the predicted values of a region-level grasp for the $i^{th}$ salient region, and $\mathcal{S}_i\in\mathbb{R}^{14\times14\times n}$ denotes the corresponding predicted segmentation. The loss of the RGPN model is defined over $\boldsymbol{I}$ salient regions as:
\begin{equation}\label{loss_rgpn}
\begin{gathered}
L_{rgpn}=\sum_{i\in\boldsymbol{I}}(L_{reg}(R_{r_i},R_{r_i}^{*})+\lambda_3L_{cls}(\theta_{r_i},\theta_{r_i}^{*})+\\
\lambda_{3}L_{cls}(\rho_{r_i},\rho_{r_i}^{*})+L_{seg}(\mathcal{S}_{i},\mathcal{S}^{*}_i)),
\end{gathered}
\end{equation}
where $R^{*}$, $\theta^{*}$, $\rho^{*}$, and $\mathcal{S}^{*}$ represent the ground truths. The term $\rho_{r_i}^{*}=0$ for a non-graspable region and $\rho_{r_i}^{*}=1$ for a graspable region. The term $L_{seg}$ represents a pixel-wise binary cross-entropy loss used to learn segmentations of salient regions. It is given by:
\begin{equation}\label{loss_srn}
L_{seg}=-\frac{1}{|\mathcal{S}_{i}|}\sum_{j\in\mathcal{S}_{i}}(y_{j}\log(\hat{y}_{j})+(1-y_{j})\log(1-\hat{y}_{j})),
\end{equation}
where, $y_{j}$ represents the ground truth value and $\hat{y}_{j}$ denotes the predicted value for a pixel $j\in\mathcal{S}_{i}$.
Learning segmentations in parallel with grasp poses helps the network to produce better localization results \cite{he2017mask}.  
The total loss of our RGN model is given by:
\begin{equation}\label{loss_rgn}
L_{rgn}=L_{srn}+L_{rgpn}.
\end{equation}
The terms $\lambda_1$, $\lambda_2$, and $\lambda_3$ in Eq. \ref{loss_ggpn}, Eq. \ref{loss_srn}, and Eq. \ref{loss_rgpn} control the relative influence of classification over regression on the combined objective functions\footnote{For experiments, we set the parameters $\lambda_1$, $\lambda_2$, and $\lambda_3$ to 0.4.}.
\subsection{Pixel Grasp network (PGN)}\label{pixel_grasp_network}
Here, we feed the features extracted from the base network into \textit{Dense block 7} followed by a group of upsampling layers which increase the spatial resolution of the features and produce feature maps of the size of the input image. These feature maps encode the parameters of the grasp pose at every pixel of the image. 
\\
\indent
Let $\boldsymbol{M}_{xy_i}$, $\boldsymbol{M}_{w_i}$, $\boldsymbol{M}_{h_i}$, and $\boldsymbol{M}_{\theta_i}$ denote the predicted feature maps of the $i^{th}$ image, respectively. We define the loss of the PGN model over $\boldsymbol{K}$ images as:
\begin{equation}\label{loss_pgn}
\begin{gathered}
L_{pgn}=\sum_{i\in \boldsymbol{K}}(L_{reg}(\boldsymbol{M}_{xy_i},\boldsymbol{M}_{xy_i}^{*})+L_{reg}(\boldsymbol{M}_{w_i},\boldsymbol{M}_{w_i}^{*})+\\
L_{reg}(\boldsymbol{M}_{h_i},\boldsymbol{M}_{h_i}^{*})+L_{cls}(\boldsymbol{M}_{\theta_i},\boldsymbol{M}_{\theta_i}^{*})),
\end{gathered}
\end{equation}
where $\boldsymbol{M}_{xy_i}^{*}$, $\boldsymbol{M}_{w_i}^{*}$, $\boldsymbol{M}_{h_i}^{*}$, and $\boldsymbol{M}_{\theta_i}^{*}$ represent the ground-truths. 
\subsection{Training and Implementation}\label{training}
For the global model, we trained the GGPN and the GEN sub-networks independently. For the region-based and the pixel-based models, we trained the networks in an end-to-end manner. Specifically, we initialized the weights of the base network with the weights pre-trained on ImageNet. For the \textit{Dense blocks (4-7)}, the fully connected layers of GGPN, GEN, SRN, and RGPN, and the fully convolutional layers of PGN, we initialized the weights from zero-mean Gaussian distributions (standard deviation set to 0.01, biases set to 0), and trained the networks using the loss functions in Eq. \ref{loss_ggpn}, Eq. \ref{loss_gen}, Eq. \ref{loss_srn}, Eq. \ref{loss_rgn}, and Eq. \ref{loss_pgn}, respectively for 150 epochs. The starting learning rate was set to 0.01 and divided by 10 at 50\% and 75\% of the total number of epochs. The parameter decay was set to 0.0005 on the weights and biases. 
\\
\indent
Our implementation is based on the framework of Torch library \cite{paszke2017automatic}. Training was performed using ADAM optimizer and data parallelism on four Nvidia Tesla K80 GPU devices. 
For grasp selection during inference, DSGD selects the most confident region-level grasp if its confidence score is greater than a confidence threshold ($\delta_{rgn}$), otherwise DSGD switches to the PGN branch and selects the most confident pixel-level grasp. If the most confident pixel-level grasp has a confidence score less than $\delta_{pgn}$, DSGD switches to the GGN branch and selects the global grasp as the output.
Experimentally, we found that $\delta_{rgn}=0.95$ and $\delta_{pgn}=0.90$ produced the best grasp detection results.
\section{Experiments}\label{experiments}
We evaluated DSGD for grasp detection on the popular Cornell grasp dataset \cite{lenz2015deep}, which contains 885 RGB-D images of 240 objects. The ground-truth is available in the form of grasp-rectangles.
We also evaluated DSGD for multi-object grasp detection in new environments. 
For this, we used the multi-object dataset of \cite{asifijcai2018} which consists of 6896 RGB-D images of indoor scenes containing multiple objects placed in different locations and orientations.
The dataset was generated using an extended version of the scene labeling framework of \cite{asif2017multi} and \cite{asif2016simultaneous}. 
For evaluation, we used the object-wise splitting criteria \cite{lenz2015deep} for both the Cornell grasp dataset and our multi-object dataset. The object-wise splitting splits the object instances randomly into train and test subsets (i.e., the training set and the test set do not share any images from the same object). This strategy evaluates how well the model generalizes to unseen objects.
For comparison purposes, we followed the procedure of \cite{redmon2015real} and substituted the blue channel with the depth image, where the depth values are normalized between 0 and 255. We also performed data augmentation through random rotations.
\\
\indent
For grasp evaluation, we used the ``rectangle-metric" proposed in \cite{jiang2011efficient}. A grasp is considered to be correct if: \textbf{i)} the difference between the predicted grasp angle and the ground-truth is less than $30^\circ$, and \textbf{ii)} the Jaccard index of the predicted grasp and the ground-truth is higher than 25\%. The Jaccard index for a predicted rectangle $\boldsymbol{\mathcal{R}}$ and a ground-truth rectangle $\boldsymbol{\mathcal{R}}^*$ is defined as:
\begin{equation}
J(\boldsymbol{\mathcal{R}}^*,\boldsymbol{\mathcal{R}})=\frac{|\boldsymbol{\mathcal{R}}^*\cap \boldsymbol{\mathcal{R}}|}{|\boldsymbol{\mathcal{R}}^*\cup \boldsymbol{\mathcal{R}}|}.
\end{equation}
\subsection{Single-Object Grasp Detection}\label{experimental_results}
\begin{table}[t!]
\caption{Grasp evaluation on the Cornell grasp dataset in terms of average grasp detection accuracy.}
\vspace{-5pt}
\setlength\tabcolsep{12pt}\centering
\begin{tabular}{@{}lcccc@{}}
\toprule
\multirow{1}{*}{Method} & \multicolumn{1}{c}{Accuracy (\%)}			\\
\midrule
(Jiang \textit{et. al.} 2011) Fast search 		&58.3 			\\
(Lenz \textit{et. al.} 2015) Deep learning      &75.6  		 	\\
(Redmon \textit{et. al.} 2015) MultiGrasp   	&87.1     		\\	
(Wang \textit{et. al.} 2015) Multi-modal   		&-     			\\
(Kumra \textit{et. al.} 2017) ResNets   		&88.9     		\\
(Guo \textit{et. al.} 2017) Hybrid-Net   		&89.1     		\\
\hline
(this work) DSGD							&\textbf{97.5}	\\
\bottomrule
\end{tabular}
\label{table_cornell}
\vspace{-5pt}
\end{table}
\begin{figure*}[t!]
  \begin{center}
    \includegraphics[trim=0.0cm 0.1cm 0.0cm 0.0cm,clip,width=1.0\linewidth,keepaspectratio]{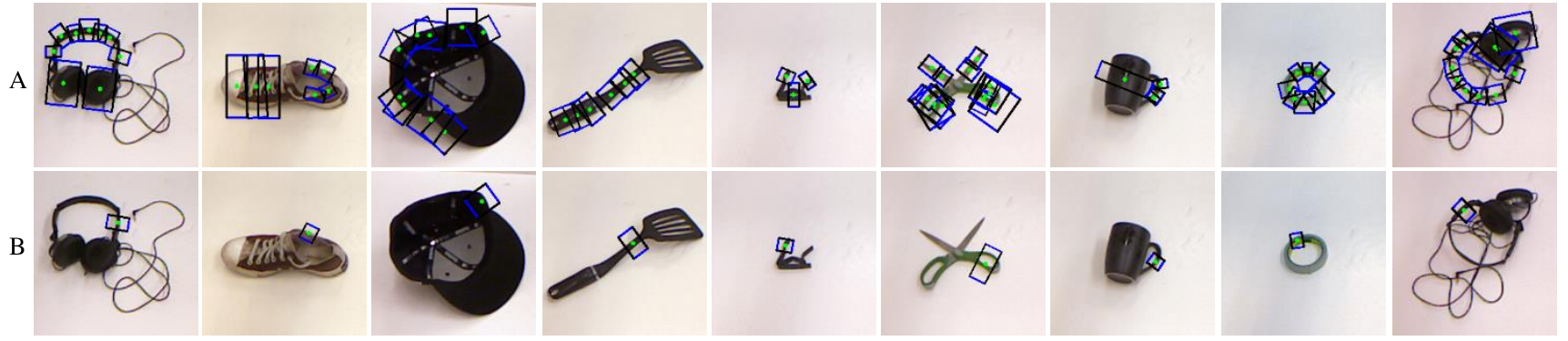}
  \vspace{-15pt}
  \caption{Grasp detection results of our DSGD (B) on some challenging objects of the Cornell grasp dataset. Ground truths are shown in A.}
  \vspace{-10pt}
  \label{fig_dsgd}
  \end{center}
\end{figure*}
Table \ref{table_cornell} shows that our DSGD achieved a considerable improvement of around 8\% in mean accuracy compared to the best performing model of \cite{guo2017hybrid} on the Cornell grasp dataset. We attribute this improvement to two main reasons: 
\textbf{First}, the proposed hierarchical grasp generation enables DSGD to produce grasps and their confidence scores from both global and local contexts. This enables DSGD to effectively recover from the errors of the global \cite{kumra2016robotic} or local methods \cite{guo2017hybrid}.
\textbf{Second}, the use of dense feature fusion enables the networks to learn more discriminative features compared to the models used in \cite{kumra2016robotic,guo2017hybrid}, respectively. 
Fig. \ref{fig_dsgd} shows grasps produced by our DSGD on some images of the Cornell grasp dataset.
\subsubsection{Significance of Combining Global and Local Models:}\label{ensemble_inference}
\begin{figure*}[t!]
  \begin{center}
    \includegraphics[trim=0.0cm 0.0cm 0.0cm 0.0cm,clip,width=1.0\linewidth,keepaspectratio]{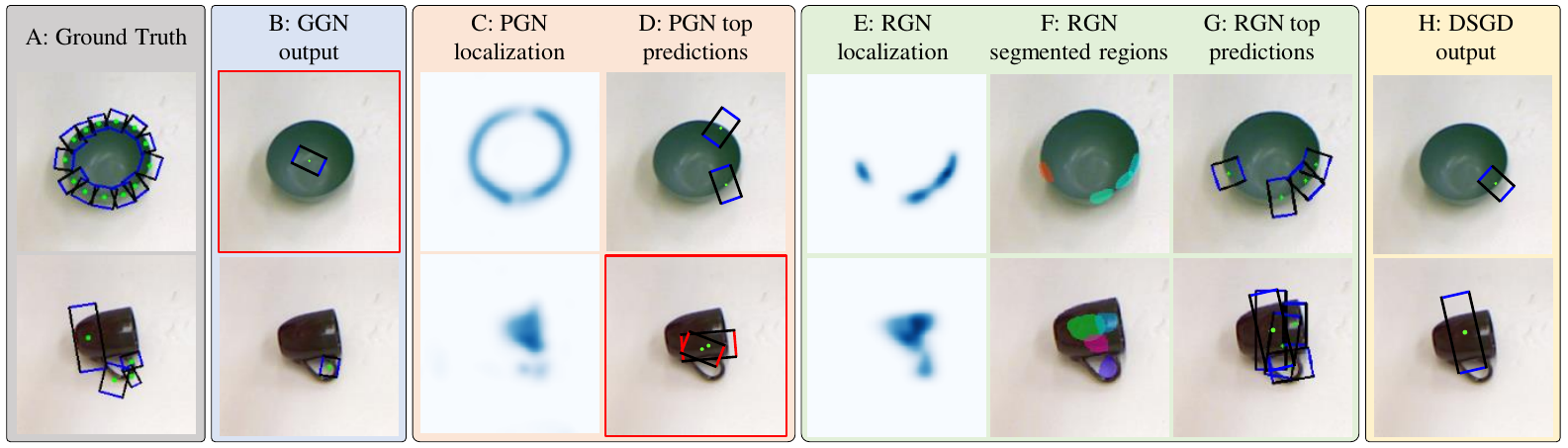}
  \vspace{-15pt}
  \caption{Qualitative comparison of grasps produced by the proposed GGN, PGN, RGN, and DSGD models. The results show that our DSGD effectively recovers from the errors of the individual models. Incorrect predictions are highlighted in red.}
  \vspace{-10pt}
  \label{fig_sgn_mgn}
  \end{center}
\end{figure*}

\begin{table}[t!]
\caption{Comparison of the individual networks of the proposed DSGD in terms of grasp accuracy (\%) on the Cornell grasp dataset.}
\vspace{-5pt}
\setlength\tabcolsep{7pt}\centering
\begin{tabular}{@{}lccccccc@{}}
\toprule
&\multirow{1}{*}{Base} 		& Global model 	&\multicolumn{2}{c}{Local models} &\multirow{2}{*}{DSGD}  \\
&network 					& {GGN} 	&PGN 	&RGN 		&	\\
\midrule
&ResNet50  		&86.8		&94.1	&96.1	&\textbf{96.7} 			   \\
&DenseNet 		&88.9		&95.4	&96.8	&\textbf{97.5}		\\
\bottomrule
\end{tabular}
\label{table_ensemble}
\vspace{-5pt}
\end{table}
Table \ref{table_ensemble} shows a quantitative comparison of the individual models of our DSGD in terms of grasp accuracy on the Cornell grasp dataset, for different CNN structures as the base network. The base networks we tested include: ResNets \cite{he2016deep} and DenseNets \cite{huang2017densely}. Table \ref{table_ensemble} shows that on average the local models (PGN and RGN) produced higher grasp accuracy compared to the global model (GGN). 
\\
\indent
The global and the local models have their own pros and cons. The global model learns an average of the ground-truth grasps from the global perspective. Although, the global-grasps are accurate for most of the objects, they tend to lie in the middle of circular symmetric objects resulting in localization errors as highlighted in red in Fig. \ref{fig_sgn_mgn}-B. The PGN model on the other hand operates at the pixel-level and produces correct grasp localizations for these challenging objects as shown in Fig. \ref{fig_sgn_mgn}-C. However, pixel-based model is susceptible to outliers in the position prediction maps which result in localization errors as highlighted in red in Fig. \ref{fig_sgn_mgn}-D.
Our RGN model works at a semi-global level while maintaining large receptive fields. It produces predictions using features extracted from the salient parts of the image which highly likely encode graspable parts of an object (e.g., handles or boundaries) as shown in Fig. \ref{fig_sgn_mgn}-F. Consequently, RGN is less susceptible to pixel-level outliers and does not suffer global averaging errors as shown in Fig. \ref{fig_sgn_mgn}-G. 
Our DSGD takes advantage of both the global context and local predictions and produces highly accurate grasps as shown in Fig. \ref{fig_sgn_mgn}-H.
%
%
%
%
%
\subsubsection{Ablative Study of the Proposed DSGD:}\label{ablation_study}
\begin{table*}[t!]
\caption{Ablation study of our DSGD (with DenseNet as the base network) on the Cornell grasp dataset in terms of the growth rate ($\mathcal{W}$) and the number of dense layers ($N_l$) of the GGN, PGN, and RGN sub-networks.}
\vspace{-5pt}
\setlength\tabcolsep{5pt}\centering
\begin{tabular}{@{}l|ccccc|ccccc|ccccc|ccccc|c@{}}
\toprule
\multirow{2}{*}{Model}& \multicolumn{5}{c|}{GGN}& \multicolumn{5}{c|}{PGN}& \multicolumn{5}{c|}{RGN}&\multirow{1}{*}{Accuracy }		 &\multirow{1}{*}{Speed }\\
& $\mathcal{W}$ 	&$N_{l_1}$&$N_{l_2}$&$N_{l_3}$&$N_{l_5}$ &$\mathcal{W}$ 	&$N_{l_1}$&$N_{l_2}$&$N_{l_3}$&$N_{l_7}$ &$\mathcal{W}$ 	&$N_{l_1}$&$N_{l_2}$&$N_{l_3}$&$N_{l_4}$  &(\%) &$(fps)$\\
\midrule
DSGD-\textit{lite}    	&32			&6&12&24&16												&32			&6&12&24&16
						&32			&6&12&24&16	&97.1	&\textbf{12}\\

DSGD-A   	 			&32  		&6&12&32&32
						&32			&6&12&32&32
						&32			&6&12&24&16 &{97.1} &11\\    						

DSGD-B   	 			&32  		&6&12&48&32
						&32			&6&12&48&32
						&32			&6&12&24&16 &{97.3} &10\\    						

DSGD-C    	 			&48  		&6&12&36&24
						&48			&6&12&36&24
						&32			&6&12&24&16	&\textbf{97.5}   &9\\
\bottomrule
\end{tabular}
\label{table_densenet}
\vspace{-5pt}
\end{table*}
\begin{figure*}[t!]
  \begin{center}
    \includegraphics[trim=0.1cm 0.1cm 0.1cm 0.1cm,clip,width=1.0\linewidth,keepaspectratio]{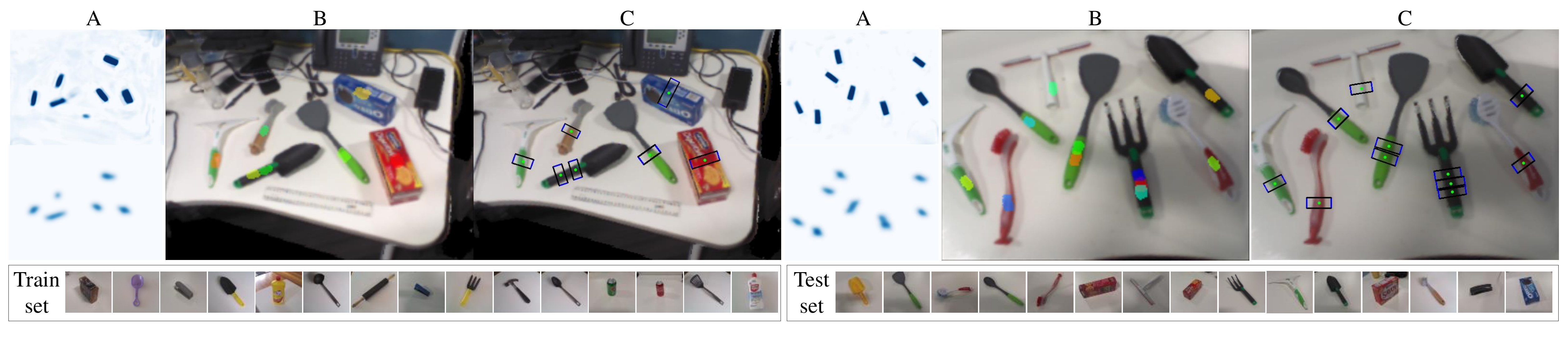}
  \vspace{-15pt}
  \caption{Grasp evaluation on our multi-object dataset. The train and the test sets do not share any images from the same object instance. The localization outputs of our pixel-level (PGN) and region-level (RGN) grasp models are shown in (A) and (B), respectively. The segmented regions produced by our RGN model are shown in (C). Note that we only show grasps with confidence scores higher than 90\% in (D).}
  \vspace{-5pt}
  \label{fig_sequence}
  \end{center}
\end{figure*}
\begin{figure*}[t!]
  \begin{center}
    \includegraphics[trim=1.4cm 0.4cm 1.4cm 0.2cm,clip,width=1.0\linewidth,keepaspectratio]{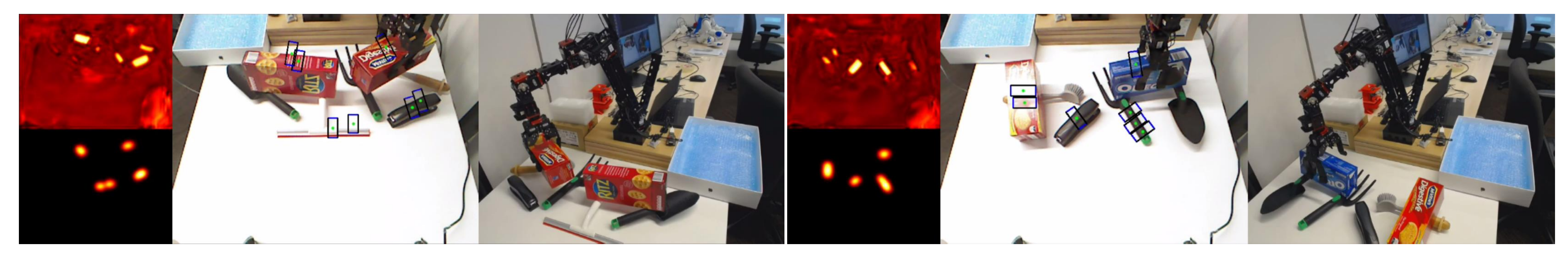}
  \vspace{-18pt}
  \caption{Experimental setting for real-world robotic grasping.}
  \vspace{-10pt}
  \label{fig_robot}
  \end{center}
\end{figure*}
The growth rate parameter $\mathcal{W}$ refers to the number of output feature maps of each dense layer and therefore controls the depth of the network. 
Table \ref{table_densenet} shows that a large growth rate and wider dense blocks (i.e., more number of layers in the dense blocks) increase the average accuracy from 97.1\% to 97.5\% at the expense of low runtime speed due to the overhead from additional channels.
Table \ref{table_densenet} also shows that a \textit{lite} version of our detector (DSGD-\textit{lite}) can run at 12\textit{fps} making it suitable for real-time applications. 
\subsection{Multi-Object Grasp Detection}\label{multi_object}
Table \ref{table_multi_object} shows our grasp evaluation on the multi-object dataset. The results show that on average, our DSGD improves grasp detection accuracy by 9\% and 2.4\% compared to the pixel-level and region-level models, respectively. 
Fig. \ref{fig_sequence} shows qualitative results on the images of our multi-object dataset. The results show that our DSGD successfully generates correct grasps for multiple objects in real-world scenes containing background clutter. 
The generalization capability of our model is attributed to the proposed hierarchical image-to-grasp mappings, where the proposed region-level network and the proposed pixel-level network learn to associate grasp poses to salient regions and salient pixels in the image data, respectively. These salient regions and pixels encode object graspable parts (e.g., boundaries, corners, handles, extrusions) which are generic (i.e., have similar appearance and structural characteristics) across a large variety of objects generally found in indoor environments. Consequently, the proposed hierarchical mappings learned by our models successfully generalize to new object instances during testing. 
This justifies the practicality of our DSGD for real-world robotic grasping.   
\subsection{Robotic Grasping}\label{robotic_grasping}
\begin{table}[t!]
\caption{Grasp evaluation on our multi-object dataset.}
\vspace{-5pt}
\setlength\tabcolsep{4pt}\centering
\begin{tabular}{@{}llccccc@{}}
\toprule
&\multirow{2}{*}{Base network}			 	&\multicolumn{3}{c}{Grasp accuracy} 				&Robotic grasp						\\
& 			 								&PGN 		&RGN 		&DSGD 				&success						\\
\midrule
&ResNet    								&86.5\%		&93.4\%		&\textbf{95.8}\%	&89\% 			   \\
&DenseNet 								&87.4\%		&94.7\%		&\textbf{97.2}\%	&90\%		\\
\bottomrule
\end{tabular}
\label{table_multi_object}
\vspace{-10pt}
\end{table}
Our robotic grasping setup consists of a Kinect for image acquisition and a 7 degrees of freedom robotic arm which is tasked to grasp, lift, and take-away the objects placed within the robot workspace. For each image, DSGD generates multiple grasp candidates as shown in Fig. \ref{fig_robot}. For grasp execution, we select a random candidate which is located within the robot workspace and has confidence greater than 90\%.
A robotic grasp is considered successful if the robot grasps a target object (verified through force sensing in the gripper), holds it in air for 3 seconds and takes it away from the robot workspace. The objects are placed in random positions and orientations to remove bias related to the object pose.
Table \ref{table_multi_object} shows the success rates computed over 200 grasping trials. The results show that we achieved grasp success rates of 90\% with DenseNet as the base network. 
Some failure cases include objects with non-planar grasping surfaces (e.g., brush). However, this can be improved by multi-finger grasps. We leave this for future work as our robotic arm only supports parallel grasps.
\section{Conclusion and Future Work}\label{conclusion}
We presented \textit{Densely Supervised Grasp Detector} (DSGD), which generates grasps and their confidence scores at different image hierarchical levels (i.e., global-, region-, and pixel-levels). 
%
%
Experiments show that our proposed hierarchical grasp generation produces superior grasp accuracy compared to the state-of-the-art on the Cornell grasp dataset.     
Our evaluations on videos from Kinect and robotic grasping experiments show the capability of our DSGD for producing stable grasps for unseen objects in new environments.
In future, we plan to reduce the computational burden of our DSGD through parameter-pruning for low-powered GPU devices.
\bibliographystyle{aaai}
\bibliography{egbib}
%
\end{document}